\documentclass{article}
\usepackage{ijcai24}

\usepackage{times}
\usepackage{soul}
\usepackage{url}
\usepackage[hidelinks]{hyperref}
\usepackage[utf8]{inputenc}
\usepackage[small]{caption}
\usepackage{graphicx}
\usepackage{amsmath}
\usepackage{amsthm}
\usepackage{booktabs}
\usepackage{algorithm}
\usepackage{algorithmic}
\usepackage[switch]{lineno}

\usepackage{amssymb,amsfonts}
\usepackage{bm}
\usepackage{multirow}
\usepackage{epstopdf}

\title{Unified  View Imputation and Feature Selection Learning for Incomplete Multi-view Data}

 \author{
 Yanyong Huang$^1$
 \and
 Zongxin Shen$^1$\and
 Tianrui Li$^2$\And
 Fengmao Lv$^2$\\
 \affiliations
 $^1$ School of Statistics, Southwestern University of Finance and Economics, Chengdu, 611130, China\\
 $^2$School of Computing and Artificial Intelligence, Southwest Jiaotong University, Chengdu, 611756, China\\
 \emails
huangyy@swufe.edu.cn,
shenzx@smail.swufe.edu.cn,
trli@swjtu.edu.cn,
fengmaolv@126.com
 }

\begin{document}
\maketitle
\begin{abstract}
Although multi-view unsupervised feature selection (MUFS) is an effective technology for reducing dimensionality in machine learning, existing methods cannot directly deal with  incomplete multi-view data where some samples are missing in certain views. These methods should first apply predetermined values to impute missing data, then perform feature selection on the complete dataset. Separating imputation and feature selection processes fails to capitalize on the potential synergy where local structural information gleaned from feature selection could guide the imputation, thereby improving the feature selection performance in turn. Additionally, previous methods only focus on leveraging  samples' local structure information, while ignoring  the intrinsic locality of the feature space. To tackle these problems, a novel MUFS method, called UNified view Imputation and Feature selectIon lEaRning (UNIFIER), is proposed. UNIFIER explores the local structure of multi-view data by adaptively learning  similarity-induced graphs from both the sample and feature spaces. Then, UNIFIER dynamically recovers the missing views, guided by the sample and feature similarity graphs during the feature selection procedure. Furthermore, the half-quadratic minimization technique is used to automatically weight different instances, alleviating the impact of outliers and unreliable restored data. Comprehensive experimental results demonstrate that UNIFIER outperforms other state-of-the-art methods.
\end{abstract}
\section{Introduction}\label{sec:introduction}
Multi-view data describes the same sample from different perspectives or forms, often represented by high-dimensional features~\cite{W.F.Liu2018,B.Zhang2021}. In practical applications, acquiring labels for multi-view data is typically difficult due to the considerable amount of time and effort required. High-dimensional, unlabeled multi-view data presents challenges for machine learning applications, such as the curse of dimensionality, performance degradation in downstream tasks, and high computational costs. The recently developed dimensionality reduction technique, multi-view unsupervised feature selection (MUFS), aims to tackle these difficulties by selecting informative features from multi-view data in an unsupervised manner~\cite{R.Zhang2019,D.Shi2023}.

The existing MUFS approaches can generally be divided into two categories. The first category of methods sequentially connects features from multiple views into a vector, and subsequently employs single-view feature selection methods such as LPscore~\cite{LapScore}, EGCFS~\cite{EGCFS}, and FSDK~\cite{FSDK}. The second category of methods directly selects features from multi-view data by considering the correlations among different views. Typically, CvLP-DCL~\cite{CvLP-DCL} learns view-specific and view-common label spaces and constructs a cross-view similarity graph to capture consensus and diversity information from multiple views. TLR-MFS~\cite{TLR-MFS} enforces a tensor low-rank constraint on the similarity graph matrix to leverage high-order consensus information among different views in the feature selection procedure. \cite{CFSMO} fully utilizes neighbor information of differing views to construct a similarity graph for feature selection. However, previous studies assumed that each sample is present across all views, which may not hold in real-world scenarios. For example, in Alzheimer's disease detection, many patients only have features from magnetic resonance imaging (MRI) and lack positron emission tomography (PET) scans due to the high cost of PET imaging~\cite{CaiSIGKDD2018}. There are few studies on unsupervised feature selection in incomplete multi-view data. \cite{CVFS} proposed a cross-view feature selection method for incomplete multi-view data. This method first fills missing samples with mean values and then uses weighted non-negative matrix factorization on the imputed data to select features. The previously mentioned MUFS methods, when applied to incomplete multi-view data, consist of two sequential stages: first, the imputation of missing data, and then the subsequent selection of features from the imputed multi-view data. However, by treating imputation and feature selection as two separate processes, these methods miss out on the potential synergy. Local structural information gleaned from feature selection could guide the imputation process and, conversely, the refined imputation could lead to improved feature selection performance. Furthermore, although these methods have captured local structural information within the samples, they have overlooked the exploration of the intrinsic local characteristics of the feature space—a crucial aspect that could enhance feature selection performance.

\begin{figure}[t]
	\centering
	\includegraphics[width=0.48\textwidth]{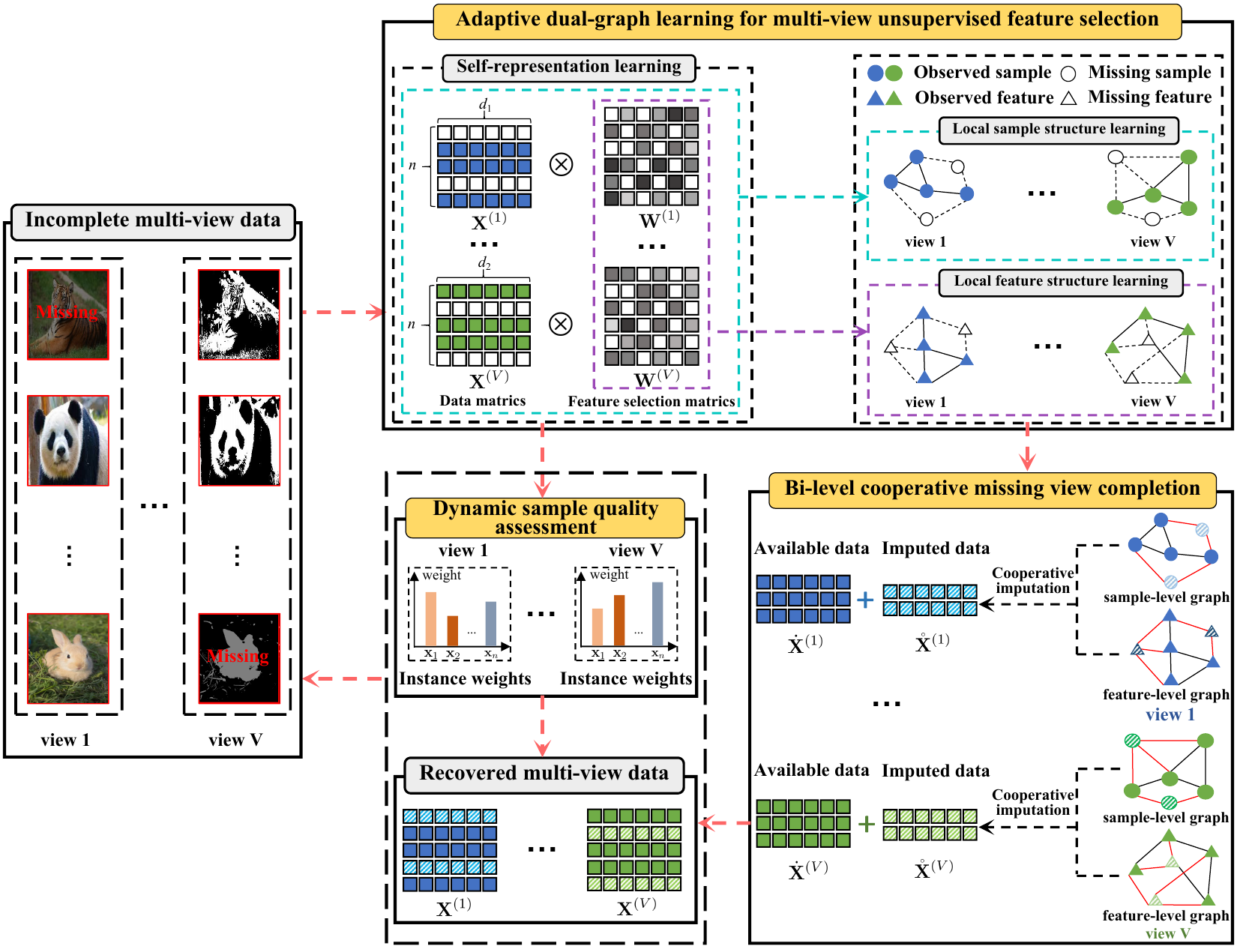}
	\caption{The framework of the proposed method UNIFIER.}\label{framework}
\end{figure}

To address the aforementioned issues, we propose a novel MUFS method named UNified view Imputation and Feature selectIon lEaRning (UNIFIER), which consists of  three key components: 1) An adaptive dual-graph learning module that selects discriminative features by leveraging the local structures of both the feature and sample spaces. 2) A bi-level cooperative missing view completion module that effectively utilizes local structure graphs at both the feature and sample levels to impute missing views in samples. 3) A dynamic sample quality assessment module that automatically weighs samples to mitigate the effects of outliers and unreliable restored data. Fig.~\ref{framework} illustrates the framework of the proposed method UNIFIER. The main contributions of this paper are  as follows:

\begin{enumerate}
	\item[\textbullet] To the best of our knowledge, this is the first work to integrate multi-view unsupervised feature selection and missing view imputation into a unified learning framework, enabling a seamless synergy between these two processes that culminates in enhanced feature selection performance.
	
	\item[\textbullet] Simultaneous investigation of the local structures in both the feature and sample spaces is conducted to facilitate missing view imputation and discriminative feature selection. Additionally, a dynamic sample quality assessment based on half-quadratic minimization is proposed, which can alleviate the impact of outliers and unreliable restored data.
	
	\item[\textbullet] An efficient alternative iterative algorithm is developed to solve the proposed UNIFIER method, and comprehensive experimental results demonstrate the superiority of UNIFIER over several state-of-the-art (SOTA) methods.
\end{enumerate}

\section{Proposed Method}\label{sec:proposed method}
We first summarize some notations throughout this paper. For any matrix $\mathbf{M}=(m_{ij}) \in \mathbb{R}^{p \times q}$, the $i$-th row and the $j$-th column  of $\mathbf{M}$ are denoted as $\mathbf{m}_{i\cdot}$ and $\mathbf{m}_{\cdot j}$, respectively. Its Frobenius norm and $\ell_{2,1}$-norm are respectively defined as  $\|\mathbf{M}\|_{F}=\sqrt{\sum_{i=1}^{p}\sum_{j=1}^{q}m_{ij}^{2}}$ and $\|\mathbf{M}\|_{2,1}=\sum_{i=1}^{p}\sqrt{\sum_{j=1}^{q}m_{ij}^{2}}$. $\operatorname{Tr}(\mathbf{M})$ and $\mathbf{M}^{\top}$ represent  the trace and transpose of $\mathbf{M}$, respectively. In the following, we will introduce three modules of the proposed method UNIFIER in detail.

\subsection{Adaptive Dual-graph Learning for MUFS}
Given a multi-view dataset $\mathcal{X}=\{\mathbf{X}^{(v)} \in \mathbb{R}^{n \times d_v}\}_{v=1}^{V}$, where $\mathbf{X}^{(v)}$ represents  the feature matrix of the $v$-th view,  $n$ and $d_v$  denote the number of samples and dimensions of $\mathbf{X}^{(v)}$, and $V$ is the number of views. In order to select discriminative features from the given multi-view dataset, a comprehensive exploration of the local structures within both the feature and sample spaces is conducted. By extending the concept of data self-representation~\cite{J.Y.LiuMM2021} to the feature level, it is reasonable to assume that each feature can be reconstructed by a linear combination of a small set of representative features. This can be formalized as follows:
\begin{equation}\label{3.2.1}
	\begin{aligned}
		&\min_{\mathbf{W}^{(v)} } \sum_{v=1}^{V} \|{\mathbf{X}^{(v)}-\mathbf{X}^{(v)}\mathbf{W}^{(v)}}\|_{F}^{2} + \lambda \|\mathbf{W}^{(v)}\|_{2,1},\\
	\end{aligned}
\end{equation}
where $\mathbf{W}^{(v)} \!\!\in\! \mathbb{R}^{d_v \times d_v}$ is the sparse representation graph or feature weight matrix in the $v$-th view. The $i$-th row of $\mathbf{W}^{v}$ indicates the contribution of the $i$-th feature $\mathbf{x}_{\cdot i}^{(v)}$ to the reconstruction of all other features, thereby reflecting the importance of the $i$-th feature in $\mathbf{X}^{v}$. Moreover, we employ the $\ell_{2,1}$-norm regularization $\|\mathbf{W}^{(v)}\|_{2,1}$ to induce sparsity within $\mathbf{W}^{v}$, which encourages the weights of less significant features to approach zero. Consequently, each feature is selectively associated with only the most representative features, effectively capturing the local structure of the feature space.

Furthermore, according to the spectral graph theory~\cite{X.W.Dong2012}, if two samples $\mathbf{x}_{i \cdot}^{(v)}$ and $\mathbf{x}_{j \cdot}^{(v)}$ are similar in high-dimensional space, they will also be similar in low-dimensional space. This allows us to preserve the local geometric structure of the sample space by adaptively learning the nearest neighborhood graph, which can be formulated as follows:
\begin{equation}\label{3.2.2}
\begin{aligned}
&\min _{\mathbf{W}^{(v)}\!, \mathbf{S}^{(v)}\!} \!\sum_{v=1}^{V} \!\sum_{i,j=1}^n \!\!\| \mathbf{x}_{i \cdot}^{(v)}\mathbf{W}^{(v)} \!-\! \mathbf{x}_{j \cdot}^{(v)}\mathbf{W}^{(v)} \!\|_{2}^{2}s_{ij}^{(v)} \!\!+\! \xi_{v} \|\mathbf{S}^{(v)} \!\|_{F}^{2}\\
&\text { s.t. } s_{ii}^{(v)}=0, s_{ij}^{(v)} \geq 0, \mathbf{1}^\top\mathbf{s}_{\cdot i}^{(v)}=1, \|\mathbf{s}_{\cdot i}^{(v)}\|_{0}=k,
\end{aligned}
\end{equation}
where $\mathbf{S}^{(v)}$ is the similarity matrix of the $v$-th view and $\xi_{v}$ is a regularization parameter. The constraint  $\|\mathbf{s}_{\cdot i}^{(v)}\|_{0}=k$ is enforced to ensure that $\mathbf{x}_{i \cdot}^{(v)}$ is exclusively connected to its $k$ nearest neighbors.

By combining Eqs. (\ref{3.2.1}) and (\ref{3.2.2}), we can obtain the following  adaptive dual-graph learning-based MUFS module:
\begin{equation}\label{3.2.3}
\begin{aligned}
&\min _{\mathbf{W}^{(v)}\!, \mathbf{S}^{(v)}\!} \sum_{v=1}^{V} \bigg[ \alpha^{(v)}\|{\mathbf{X}^{(v)}\!-\!\mathbf{X}^{(v)}\mathbf{W}^{(v)}}\|_{F}^{2} + \lambda \|\mathbf{W}^{(v)}\|_{2,1}\\
&+\frac{1}{2}\sum_{i,j=1}^n \| \mathbf{x}_{i \cdot}^{(v)}\mathbf{W}^{(v)}-\mathbf{x}_{j \cdot}^{(v)}\mathbf{W}^{(v)}\|_{2}^{2}s_{ij}^{(v)} 
+ \xi_{v} \|\mathbf{S}^{(v)} \|_{F}^{2} \bigg]\\
&\text { s.t. } s_{ii}^{(v)}=0, s_{ij}^{(v)} \geq 0, \mathbf{1}^\top\mathbf{s}_{\cdot i}^{(v)}=1, \|\mathbf{s}_{\cdot i}^{(v)}\|_{0}=k,
\end{aligned}
\end{equation}
where $\alpha^{(v)}$ is the view weight. Eq. (\ref{3.2.3}) can simultaneously use the local structure information of both the feature and sample spaces to improve the feature selection performance. 

\subsection{Bi-level Cooperative Missing View Completion}
In the context of an incomplete multi-view scenario, where some samples are present in certain views but absent in others, we use $\mathring{\mathbf{X}}^{(v)} \in \mathbb{R}^{m\times d_{v}}$ and $\dot{\mathbf{X}}^{(v)} \in \mathbb{R}^{(n-m)\times d_{v}}$ to represent the feature matrices of missing samples and available samples in $\mathbf{X}^{(v)}$, respectively. Unlike these two-step methods, which sequentially address missing value imputation followed by feature selection, we integrate feature selection and missing view imputation into a unified optimization process. To this end, we incorporate the missing data as a variable into the learning process described by Eq. (\ref{3.2.3}), which will undergo alternate optimization with other variables until convergence. The corresponding objective function is presented  below.
\begin{equation}\label{3.3.2}
\begin{aligned}
&\min _{\mathbf{W}^{(v)}\!,\mathring{\mathbf{X}}^{(v)}\!, \mathbf{S}^{(v)}\!} \sum_{v=1}^{V} \!\bigg[ \alpha^{(v)}\|{\widetilde{\mathbf{X}}^{(v)}\!-\!\widetilde{\mathbf{X}}^{(v)}\mathbf{W}^{(v)}}\|_{F}^{2} \!+\! \lambda \|\mathbf{W}^{(v)}\|_{2,1}\\
&+\frac{1}{2}\sum_{i,j=1}^n \| \widetilde{\mathbf{x}}_{i \cdot}^{(v)}\mathbf{W}^{(v)}-\widetilde{\mathbf{x}}_{j \cdot}^{(v)}\mathbf{W}^{(v)}\|_{2}^{2}s_{ij}^{(v)} 
+ \xi_{v} \| \mathbf{S}^{(v)} \|_{F}^{2} \bigg]\\
&\text { s.t. } \widetilde{\mathbf{X}}^{(v)} = \dot{\mathbf{K}}^{(v)}\dot{\mathbf{X}}^{(v)} + \mathring{\mathbf{K}}^{(v)}\mathring{\mathbf{X}}^{(v)}, s_{ii}^{(v)}=0, s_{ij}^{(v)} \geq 0, \\
& \qquad \mathbf{1}^\top\mathbf{s}_{\cdot i}^{(v)}=1, \|\mathbf{s}_{\cdot i}^{(v)}\|_{0}=k,
\end{aligned}
\end{equation}
where $\mathring{\mathbf{K}}^{(v)}\in \mathbb{R}^{n \times m}$ and $\dot{\mathbf{K}}^{(v)}\in \mathbb{R}^{n \times (n-m)}$ serve as two indicator matrices. These matrices are used to project the learned missing samples and available samples into a complete multi-view data $\widetilde{\mathbf{X}}^{(v)}$, and their specific definitions are as follows:
\begin{equation}\label{3.3.1}
\begin{aligned}
\dot{\mathbf{K}}^{(v)}_{ij} & = \begin{cases}1, & \text { if } \mathbf{x}_{i \cdot}{(v)} \text { corresponds to }  \dot{\mathbf{x}}_{j \cdot}^{(v)}\\
0, & \text { otherwise. }\end{cases} \\
\mathring{\mathbf{K}}^{(v)}_{ij} & = \begin{cases}1, & \text { if } \mathbf{x}_{i \cdot}^{(v)}  \text { corresponds to } \mathring{\mathbf{x}}_{j \cdot}^{(v)}\\
0, & \text { otherwise. }\end{cases} \\
\end{aligned}
\end{equation}

In Eq. (\ref{3.3.2}), the missing data $\mathring{\mathbf{X}}^{(v)}$ is imputed with collaborative guidance from the sample-level and feature-level graphs obtained from Eq. (\ref{3.2.3}). This imputation, in turn, enhances the learning of local both sample and feature structures. These two processes boost each other in an interplay manner to recover incomplete multi-view data and achieve better feature selection performance.

\subsection{Dynamic Sample Quality Assessment}
Due to the sensitivity of the Frobenius norm-based loss function to outliers~\cite{NieF2010NIPS}, we adopt the Geman-McClure estimator $\mathcal{L}(z) = \frac{\gamma z^2}{\gamma + z^2}$ ($\gamma$ is the scale parameter)~\cite{BarronCVPR2019} as the loss function in Eq. (\ref{3.3.2}), aiming to reduce the influence of outliers. However, the objective function based on the Geman-McClure loss function is difficult to solve. To cope with this, we replace $\mathcal{L}(z)$ with  the following equivalent expression according to the half-quadratic minimization theory~\cite{Nikolova2007}.
\begin{equation}
    \begin{aligned}
        \min _{z} \mathcal{L}(z) = \frac{\gamma z^2}{\gamma+ z^2} \Longleftrightarrow \min _{z,e} \{ ez^2 + \psi(e) \}
    \end{aligned}
\end{equation}
where $e$ is an auxiliary variable, and $\psi(e)=\gamma(\sqrt{e}-1)^{2}$ is the dual potential function of Geman-McClure loss function.

Then, the final objective function of the proposed UNIFIER is formulated as 
\begin{equation}\label{3.3.3}
\begin{aligned}
&\min_{\varTheta}\!\! \sum_{v=1}^{V} \!\!\bigg[ \!\alpha^{(v)} \!\!\sum_{i=1}^{n} \!\left( \!\!e_{i}^{(v)}\| \widetilde{\mathbf{x}}_{i \cdot}^{(v)} \!-\! \widetilde{\mathbf{x}}_{i \cdot}^{(v)}\mathbf{W}^{(v)} \|_{2}^{2} \!+\! \gamma_{v}(\sqrt{e_{i}^{(v)}} \!-\! 1)^{2}\!\!\right)\\
& + \frac{1}{2}\sum_{i,j=1}^n \| \widetilde{\mathbf{x}}_{i \cdot}^{(v)} \mathbf{W}^{(v)} -\widetilde{\mathbf{x}}_{j \cdot}^{(v)}\mathbf{W}^{(v)} \|_{2}^{2}s_{ij}^{(v)} + \xi_{v}\sum_{i=1}^{n} \| \mathbf{s}_{\cdot i}^{(v)} \|_{2}^{2}\\
& +\lambda\|\mathbf{W}^{(v)}\|_{2,1}\bigg]\\
&\text { s.t. } \widetilde{\mathbf{X}}^{(v)} = \dot{\mathbf{K}}^{(v)}\dot{\mathbf{X}}^{(v)} + \mathring{\mathbf{K}}^{(v)}\mathring{\mathbf{X}}^{(v)}, s_{ii}^{(v)}=0, s_{ij}^{(v)} \geq 0, \\
& \qquad \mathbf{1}^\top\mathbf{s}_{\cdot i}^{(v)}=1, \|\mathbf{s}_{\cdot i}^{(v)}\|_{0}=k,
\end{aligned}
\end{equation}
where $ \varTheta \!\!\!=\!\!\!\{ \mathbf{W}^{(v)},\mathring{\mathbf{X}}^{(v)},\mathbf{S}^{(v)},\mathbf{e}^{(v)} | v=1\dots n \}$, and $\mathbf{e}^{(v)} = [e_{1}^{(v)},e_{2}^{(v)},\dots,e_{n}^{(v)}]^{T} \in \mathbb{R}^{n \times 1}$. 

In Eq.~(\ref{3.3.3}), $e_{i}^{(v)}$ can assess  the quality of the $i$-th sample $\widetilde{\mathbf{x}}_{i \cdot}^{(v)}$ based on the optimization result in Eq.~(\ref{4.17}). Specifically,  if $\widetilde{\mathbf{x}}_{i \cdot}^{(v)}$  is an outlier or an unreliable restored sample, the reconstruction error $\| \widetilde{\mathbf{x}}_{i \cdot}^{(v)} - \widetilde{\mathbf{x}}_{i \cdot}^{(v)}\mathbf{W}^{(v)} \|_{2}^{2}$ will be large. As a result, the corresponding weight of $\widetilde{\mathbf{x}}_{i \cdot}^{(v)}$ will be small, and vice versa. Furthermore, $e_{i}^{(v)}$ is updated adaptively, enabling the automatic assignment of weights to samples of varying qualities.

\section{Optimization}\label{sec:optimization}
Since Eq. (\ref{3.3.3}) is not jointly convex to all four groups of variables, including  $\mathbf{W}^{(v)},\mathring{\mathbf{X}}^{(v)},\mathbf{S}^{(v)}$ and $\mathbf{e}^{(v)}$, we propose to optimize them alternatively, i.e., by fixing three groups of variables and optimizing the remaining one alternately.

\noindent\textbf{Update $\mathbf{W}^{(v)}$ by Fixing Others.} When other variables are fixed, the objective function w.r.t. $\mathbf{W}^{(v)}$ becomes:
\begin{equation}\label{4.1}
\begin{aligned}
\min_{\mathbf{W}^{(v)}} &\alpha^{(v)} \sum_{i=1}^{n} e_{i}^{(v)}\| \widetilde{\mathbf{x}}_{i \cdot}^{(v)} - \widetilde{\mathbf{x}}_{i \cdot}^{(v)}\mathbf{W}^{(v)} \|_{2}^{2} + \lambda\|\mathbf{W}^{(v)}\|_{2,1} \\
&+ \frac{1}{2}\sum_{i,j=1}^n \| \widetilde{\mathbf{x}}_{i \cdot}^{(v)} \mathbf{W}^{(v)} -\widetilde{\mathbf{x}}_{j \cdot}^{(v)}\mathbf{W}^{(v)} \|_{2}^{2}s_{ij}^{(v)} 
\end{aligned}
\end{equation}

By using the matrix trace property, we can optimize  problem (\ref{4.1}) by solving its equivalent form:
\begin{equation}\label{4.3}
\begin{aligned}
\min_{\mathbf{W}^{(v)}}& \alpha^{(v)} \!\| \mathbf{E}^{(v)}\!(\widetilde{\mathbf{X}}^{(v)}\!\!-\!\!\widetilde{\mathbf{X}}^{(v)}\mathbf{W}^{(v)})\!\|_{F}^{2} \!\!+\! \lambda\! \operatorname{Tr}(\mathbf{W}^{(v)^\top}\!\mathbf{D}^{(v)}\mathbf{W}^{(v)}) \\
&+ \operatorname{Tr}\left( \mathbf{W}^{(v)^\top} \widetilde{\mathbf{X}}^{(v) ^\top} \mathbf{L}^{(v)} \widetilde{\mathbf{X}}^{(v)} \mathbf{W}^{(v)} \right)
\end{aligned}
\end{equation}
where $\mathbf{D}^{(v)}$ is a diagonal matrix with its $i$-th diagonal entry given by  $\mathbf{D}^{(v)}(i,i)= 1 / 2\sqrt{\|\mathbf{w}_{i \cdot}^{(v)}\|_{2}^{2}+\epsilon}$ ( $\epsilon$ is a small constant to prevent the denominator from vanishing), $\mathbf{L}^{(v)}= \mathbf{G}^{(v)}-\mathbf{S}^{(v)}$ is the Laplacian matrix, and $\mathbf{E}^{(v)}=diag(\sqrt{\mathbf{e}^{(v)}})$.

By taking the derivative of Eq. (\ref{4.3}) w.r.t. $\mathbf{W}^{(v)}$ and setting it to zero, we can obtain the solution of $\mathbf{W}^{(v)}$ as follows:
\begin{equation}\label{4.5}
\mathbf{W}^{(v)}=\alpha^{(v)}\left(\mathbf{C}^{(v)}+\lambda \mathbf{D}^{(v)}\right)^{-1}\widetilde{\mathbf{X}}^{(v) ^\top}\mathbf{E}^{(v)}\mathbf{E}^{(v)^\top}\widetilde{\mathbf{X}}^{(v)}
\end{equation}
where $\mathbf{C}^{(v)}=\alpha^{(v)}\widetilde{\mathbf{X}}^{(v) ^\top}\mathbf{E}^{(v)}\mathbf{E}^{(v)^\top}\widetilde{\mathbf{X}}^{(v)} + \widetilde{\mathbf{X}}^{(v) ^\top}\mathbf{L}^{(v)}\widetilde{\mathbf{X}}^{(v)}$.
\begin{algorithm}[t]
\caption{Iterative Algorithm of UNIFIER}
\label{alg:algorithm}
\textbf{Input}:Incomplete multi-view data $\{\mathbf{X}^{(v)}\!\in\! \mathbb{R}^{n \times d_v}\}_{v=1}^{V}$; the parameters $\alpha^{(v)}$ and $\lambda$; the number of selected features $h$.

\begin{algorithmic}[1]
    \STATE Initialize $\mathbf{D}^{(v)}$, $\mathbf{e}^{(v)}$, $\mathring{\mathbf{X}}^{(v)}$ and $\mathbf{S}^{(v)}$ ($v=1,\dots,V$).
    \WHILE{not convergent}

\STATE Update $\{\mathbf{W}^{(v)}\}_{v=1}^{V}$ via Eq. (\ref{4.5});\\
\STATE Update $\{\mathbf{D}^{(v)}(i,i)= 1 / (2\sqrt{\|\mathbf{w}_{i \cdot}^{(v)}\|_{2}^{2}+\epsilon})\}_{v=1}^{V}$;\\
\STATE Update $\{\mathring{\mathbf{X}}^{(v)}\}_{v=1}^{V}$ by solving Eq. (\ref{4.9});\\
\STATE Update $\{\mathbf{S}^{(v)}\}_{v=1}^{V}$ via Eq. (\ref{4.14});\\
\STATE Update $\{\mathbf{e}^{(v)}\}_{v=1}^{V}$ via Eq. (\ref{4.17});\\
\ENDWHILE
\end{algorithmic}
\textbf{Output}:{Sorting the $\ell_{2}$-norm of the rows of $\{\mathbf{W}^{(v)} \}_{v=1}^{V}$ in descending order and selecting the top $h$ features from ${\mathbf{X}}^{(v)}$.}
\end{algorithm}

\noindent\textbf{Update $\mathring{\mathbf{X}}^{(v)}$ by Fixing Others.} After fixing the other variables, the objective function w.r.t. $\mathring{\mathbf{X}}^{(v)}$  is reduced to:
\begin{equation}\label{4.6}
\begin{aligned}
&\min_{\mathring{\mathbf{X}}^{(v)}} \alpha^{(v)} \sum_{i=1}^{n} e_{i}^{(v)}\| \widetilde{\mathbf{x}}_{i \cdot}^{(v)} - \widetilde{\mathbf{x}}_{i \cdot}^{(v)}\mathbf{W}^{(v)} \|_{2}^{2}\\
&+ \frac{1}{2}\sum_{i,j=1}^n \| \widetilde{\mathbf{x}}_{i \cdot}^{(v)} \mathbf{W}^{(v)} -\widetilde{\mathbf{x}}_{j \cdot}^{(v)}\mathbf{W}^{(v)} \|_{2}^{2}s_{ij}^{(v)}\\
&\text { s.t. } \widetilde{\mathbf{X}}^{(v)} = \dot{\mathbf{K}}^{(v)}\dot{\mathbf{X}}^{(v)} + \mathring{\mathbf{K}}^{(v)}\mathring{\mathbf{X}}^{(v)}
\end{aligned}
\end{equation}

Incorporating the equality constraint $\widetilde{\mathbf{X}}^{(v)} = \dot{\mathbf{K}}^{(v)}\dot{\mathbf{X}}^{(v)} + \mathring{\mathbf{K}}^{(v)}\mathring{\mathbf{X}}^{(v)}$ into the objective function and removing the irrelevant terms w.r.t $\mathring{\mathbf{X}}^{(v)}$, we can reformulate problem (\ref{4.6}) as follows:

\begin{equation}\label{4.8}
\begin{aligned}
&\min_{\mathring{\mathbf{X}}^{(v)}}\! \operatorname{Tr} \{\alpha^{(v)}[2\mathbf{R}^{(v) ^\top}\!\mathring{\mathbf{X}}^{(v)}\! (\mathbf{I} \!+\! \mathbf{W}^{(v)}\mathbf{W}^{(v) ^\top} \!\!-\! \mathbf{W}^{(v)} \!-\! \mathbf{W}^{(v) ^\top}\!) \\
&+\! \mathring{\mathbf{X}}^{(v) ^\top}\!\mathring{\mathbf{K}}^{(v) ^\top}\!\mathbf{H}^{(v)} \mathring{\mathbf{K}}^{(v)}\mathring{\mathbf{X}}^{(v)}(\mathbf{I} \!+\! \mathbf{W}^{(v)}\mathbf{W}^{(v) ^\top} \!\!-\! 2\mathbf{W}^{(v)}) ]+\\
& \mathbf{W}^{(v) ^\top}\!( 2\dot{\mathbf{X}}^{(v) ^\top}\!\dot{\mathbf{K}}^{(v) ^\top} \!\!+ \mathring{\mathbf{X}}^{(v) ^\top}\mathring{\mathbf{K}}^{(v) ^\top})\mathbf{L}^{(v)}\mathring{\mathbf{K}}^{(v)}\mathring{\mathbf{X}}^{(v)}\mathbf{W}^{(v)} \},\\
\end{aligned}
\end{equation}
where $\mathbf{H}^{(v)} = \mathbf{E}^{(v)}\mathbf{E}^{(v) ^\top}$, $\mathbf{R}^{(v)} = \mathring{\mathbf{K}}^{(v) ^\top}\mathbf{H}^{(v)}\dot{\mathbf{K}}^{(v)}\dot{\mathbf{X}}^{(v)}$.

Taking the derivative of Eq. (\ref{4.8}) w.r.t. $\mathring{\mathbf{X}}^{(v)}$ and setting it to zero, we can obtain:
\begin{equation}\label{4.9}
\begin{aligned}
\mathbf{A}^{(v)}\mathring{\mathbf{X}}^{(v)}\mathbf{W}^{(v)}\mathbf{W}^{(v) ^\top} \!+ \mathbf{P}^{(v)}\mathring{\mathbf{X}}^{(v)}\mathbf{Q}^{(v)}=\mathbf{F}^{(v)},
\end{aligned}
\end{equation}
where $\mathbf{A}^{(v)} = \mathring{\mathbf{K}}^{(v) ^\top}\mathbf{L}^{(v)}\mathring{\mathbf{K}}^{(v)}$, $\mathbf{Q}^{(v)}=\mathbf{I}-\mathbf{W}^{(v)}-\mathbf{W}^{(v) ^\top}+\mathbf{W}^{(v)}\mathbf{W}^{(v) ^\top}$, $\mathbf{P}^{(v)}=\alpha^{(v)}\mathring{\mathbf{K}}^{(v) ^\top}\mathbf{H}^{(v)}\mathring{\mathbf{K}}^{(v)}$, $\mathbf{F}^{(v)}=\alpha^{(v)}\mathbf{R}^{(v)}\mathbf{Q}^{(v)}+\mathring{\mathbf{K}}^{(v) ^\top}\mathbf{L}^{(v)}\dot{\mathbf{K}}^{(v)}\dot{\mathbf{X}}^{(v)}\mathbf{W}^{(v)}\mathbf{W}^{(v) ^\top}$. Problem (\ref{4.9}) is a generalized Sylvester matrix equation, which can be solved according to the method proposed in~\cite{Zhang2017slove}.

\noindent\textbf{Update $\mathbf{S}^{(v)}$ by Fixing Others.} By fixing the other variables, the optimization problem w.r.t $\mathbf{S}^{(v)}$ in Eq. (\ref{3.3.3}) becomes independent for different $i$, allowing us to optimize $\mathbf{S}^{(v)}$ by separately solving each $\mathbf{s}_{\cdot i}^{(v)}$ as follows:
\begin{equation}\label{4.12}
\begin{aligned}
&\min_{\mathbf{s}_{\cdot i}^{(v)}} \frac{1}{2}\sum_{j=1}^{n}\| \widetilde{\mathbf{x}}_{i \cdot}^{(v)} \mathbf{W}^{(v)} -\widetilde{\mathbf{x}}_{j \cdot}^{(v)}\mathbf{W}^{(v)} \|_{2}^{2}s_{ij}^{(v)} + \xi_{v} \left\| \mathbf{s}_{\cdot i}^{(v)} \right\|_{2}^{2} \\
&\text { s.t. } s_{ii}^{(v)}=0,s_{ij}^{(v)} \geq 0, \mathbf{1}^{\top}\mathbf{s}_{\cdot i}^{(v)}=1, \|\mathbf{s}_{\cdot i}^{(v)}\|_{0}=k,
\end{aligned}
\end{equation}

By defining a vector $\mathbf{b}_{i}^{(v)}$ with the $j$-th entry as $b_{ij}^{(v)}=\frac{1}{2}\| \widetilde{\mathbf{x}}_{i \cdot}^{(v)} \mathbf{W}^{(v)} -\widetilde{\mathbf{x}}_{j \cdot}^{(v)}\mathbf{W}^{(v)} \|_{2}^{2}$, problem (\ref{4.12}) can be transformed into:
\begin{equation}\label{4.13}
\begin{aligned}
&\min _{\mathbf{s}_{\cdot i}^{(v)}} \frac{1}{2}\left\|\mathbf{s}_{\cdot i}^{(v)}+ \mathbf{b}_{i}^{(v)} / 2 \xi_{v} \right\|_{2}^{2} \\
\text { s.t. } s_{i i}^{(v)}=0, &s_{i j}^{(v)} \geq 0, \mathbf{1}^{\top} \mathbf{s}_{\cdot i}^{(v)}=1, \|\mathbf{s}_{\cdot i}^{(v)}\|_{0}=k.
\end{aligned}
\end{equation}

Then, following the same derivation process in~\cite{NieTKDE2019}, we can get the optimal solution of $s_{ij}^{(v)}$ as follows:
\begin{equation}\label{4.14}
s_{i j}^{(v)}=\left\{\begin{array}{cc}
\frac{b_{i, k+1}^{(v)}-b_{i j}^{(v)}}{k b_{i, k+1}^{(v)}-\sum_{t=1}^{k} b_{i t}^{(v)}} & j \leq k \\
0 & j>k
\end{array}\right.
\end{equation}

Besides, the regularization parameter $\xi_{v}$ is determined as $(kb_{i,k+1}^{(v)}-\sum_{t=1}^{k}b_{it}^{(v)})/2$ to ensure that $\mathbf{s}_{\cdot i}^{(v)}$ only has $k$ nonzero entries.

\noindent\textbf{Update $\mathbf{e}^{(v)}$ by Fixing Others.} When other variables are fixed, the  objective function w.r.t. $\mathbf{e}^{(v)}$ becomes:
\begin{equation}\label{4.15}
\min_{\mathbf{e}^{(v)}}\sum_{i=1}^{n} \left( e_{i}^{(v)}\| \widetilde{\mathbf{x}}_{i \cdot}^{(v)} - \widetilde{\mathbf{x}}_{i \cdot}^{(v)}\mathbf{W}^{(v)} \|_{2}^{2} + \gamma_{v}(\sqrt{e_{i}^{(v)}}-1)^{2}\right) 
\end{equation}

By taking the derivative of Eq. (\ref{4.15}) w.r.t. $e_{i}^{(v)}$ and setting it to zero, we can get the solution of $e_{i}^{(v)}$ as follows:
\begin{equation}\label{4.17}
e_{i}^{(v)} = \left( \frac{\gamma_{v}}{ \gamma_{v} + \| \widetilde{\mathbf{x}}_{i \cdot}^{(v)} - \widetilde{\mathbf{x}}_{i \cdot}^{(v)}\mathbf{W}^{(v)} \|_{2}^{2} } \right)^{2}
\end{equation}

Algorithm 1 summarizes the detailed steps to solve Eq. (\ref{3.3.3}). In Algorithm 1, $\mathbf{D}^{(v)}$ is initialized as an identity matrix,  ${e}_{i}^{(v)}$ is set as $1/n$ for all views, $\{\mathring{\mathbf{X}}^{(v)}\}_{v=1}^{l}$ is initialized by the mean values of $\dot{\mathbf{X}}^{(v)}$, and  $\mathbf{S}^{(v)}$ is initialized by constructing the $k$-nearest neighbor graph according to~\cite{X.L.Li2019}.

\section{Algorithm Analysis}\label{sec:Discussions}
\textbf{Time complexity analysis.} In Algorithm 1, the updates for $\mathbf{W}^{(v)}$, $\mathring{\mathbf{X}}^{(v)}$, $\mathbf{S}^{(v)}$ and $\mathbf{e}^{v}$ are performed alternately. In each iteration, updating $\mathbf{W}^{(v)}$ costs $\mathcal{O}(d_{v}^{3} + \operatorname{max}(n,d_{v})nd_{v})$ and updating $\mathring{\mathbf{X}}^{(v)}$ costs $\mathcal{O}( (\operatorname{max}(m,d_{v})md_{v})t )$, where $t$ denotes the iteration number of conjugate gradient descent. The updates of $\mathbf{S}^{(v)}$ and $\mathbf{e}^{v}$ only involve element-based operations, thus their computational cost can be ignored. In summary, the time complexity of each iteration in Algorithm 1 is $\mathcal{O}(V(\operatorname{max}(n,d_{v})nd_{v}+(\operatorname{max}(m,d_{v})md_{v})t+d_{v}^{3}))$.

\noindent\textbf{Convergence analysis.}
Following the similar derivation in \cite{C.P.HouConv}, it is easy to demonstrate that  updating $\mathbf{W}^{(v)}$ by solving problem (\ref{4.3}) leads to a monotonic decrease in the objective value of Eq. (\ref{3.3.3}). Besides, the convergence of optimizing  $\mathring{\mathbf{X}}^{(v)}$ can be guaranteed according to \cite{Zhang2017slove}. Moreover,  the closed-form solutions of Eqs.~(\ref{4.14}) and (\ref{4.17})  demonstrate the convergence of the updates for $\mathbf{S}^{(v)}$ and $\mathbf{e}^{(v)}$. Thus, the objective function value of Eq. (\ref{3.3.3}) will monotonically decrease in each iteration until convergence. 

\section{Experiments}\label{sec:Experimental}
\subsection{Experimental Settings}
\textbf{Datasets.}
We evaluate the performance of the proposed UNIFIER on six real-world multi-view datasets, including two text datasets: BBCSport~\cite{BBCS} and BBC4views~\cite{BBC4}; a face image dataset: Yale~\cite{Yale}; two object image datasets: Aloi~\cite{Aloi} and Caltech101-20~\cite{Cal20}; and  a handwritten digit image dataset: USPS2View~\cite{USPS}. A detailed description of the datasets is summarized in Table~\ref{Table1}. To simulate the incomplete multi-view setting, we follow the method in~\cite{LinYTPAMI2022} by randomly removing $\lfloor n \times r \rfloor$ instances from each view, with $r$ representing the missing data ratio and $\lfloor \cdot \rfloor$ denoting the round-down operator. In the experiment, we vary the value of  $r$ over the range of  $\{0.1, 0.15, 0.2, 0.25, 0.3\}$.

\begin{table}[!htbp]
\tabcolsep 0pt
\caption{Dataset description}\label{Table1}
\vspace*{-0.4cm}
\renewcommand\tabcolsep{1.5pt} 
    \begin{flushleft}
    \def\temptablewidth{\textwidth}
        \resizebox{8.5cm}{!}{
        \begin{tabular}{@{\extracolsep{\fill}}lccccc}
            \toprule
            Datasets  &Abbr.& Views& Instances & Features & Classes\\
            \hline
            BBCSport& BBCS       & 4 & 116 &1991/2063/2113/2158 &  5\\

            Yale &Yale     & 3 & 165  & 4096/3304/6750 & 15 \\

			BBC4views& BBC4        &4   &685  &4659/4633/4665/4684  &5\\
			
			Aloi &Aloi         & 4 &  1869&  77/13/64/64&  17  \\

            Caltech101-20&Cal20         & 6 &  2386&  48/40/254/1984/512/928 &  20 \\

            USPS2View &USPS   &  2 &   5427 &    256/32 &  5 \\
              \hline
        \end{tabular}}
    \end{flushleft}
\end{table}

\begin{table*}[t]
\centering
\caption{Means (\%) of ACC and NMI of different methods on six datasets with missing ratio 0.2 while selecting 40\% of all features.}\label{Table2}
\vspace{-0.2cm}
\resizebox{\textwidth}{!}{
\renewcommand\tabcolsep{5pt}
\small
\begin{tabular}{lllllllllllllll}
\toprule[1pt]
\multirow{2}*{Methods} & \multicolumn{2}{c}{BBCS} & \multicolumn{2}{c}{Yale} & \multicolumn{2}{c}{BBC4} & \multicolumn{2}{c}{Aloi} & \multicolumn{2}{c}{Cal20} & \multicolumn{2}{c}{USPS} \\
\cmidrule(r){2-3} \cmidrule(r){4-5} \cmidrule(r){6-7} \cmidrule(r){8-9} \cmidrule(r){10-11} \cmidrule(r){12-13}
~& ACC & NMI & ACC & NMI & ACC & NMI & ACC & NMI & ACC & NMI & ACC & NMI \\
\hline
UNIFIER & \textbf{71.55} & \textbf{54.26} & \textbf{61.21} & \textbf{59.66} & \textbf{68.76} & \textbf{47.45} & \textbf{74.91} & \textbf{78.74} & \textbf{67.48} & \textbf{55.32} & \textbf{62.87} & \textbf{41.93} \\
AllFea & 31.90 & 5.47 & 38.79 & 40.96 & 33.14 & 5.80 & 43.02 & 47.01 & 64.67 & 51.63 & 27.55 & 3.69 \\
LPscore & 37.07 & 12.96 & 48.48 & 52.09 & 44.67 & 11.24 & 17.98 & 20.31 & 47.57 & 37.00 & 27.31 & 6.79 \\
EGCFS & 49.14 & 28.86 & 33.33 & 38.27 & 37.08 & 3.73 & 17.28 & 14.43 & 41.16 & 27.33 & 27.40 & 7.12 \\
HMUFS & 56.90 & 30.89 & 52.73 & 53.48 & 42.63 & 16.88 & 68.33 & 72.50 & 65.47 & 53.99 & 57.01 & 35.20 \\
FSDK & 65.00 & 51.39 & 51.30 & 51.31 & 47.63 & 31.02 & 68.35 & 70.27 & 65.72 & 51.92 & 55.92 & 33.72 \\
NSGL & 58.62 & 34.00 & 56.97 & 55.64 & 42.63 & 17.06 & 61.16 & 63.59 & 65.21 & 52.45 & 57.71 & 37.68 \\
CVFS & 55.17 & 31.12 & 55.76 & 54.98 & 32.99 & 18.75 & 47.87 & 19.33 & 63.79 & 50.34 & 56.13 & 36.14 \\
CvLP-DCL & 51.72 & 26.24 & 49.09 & 52.39 & 33.14 & 16.39 & 63.62 & 67.66 & 65.93 & 51.36 & 58.95 & 37.96 \\
TLR-MFS & 67.24 & 49.97 & 52.12 & 52.21 & 55.47 & 30.16 & 69.40 & 71.00 & 65.13 & 50.67 & 61.27 & 39.69 \\
JMVFG & 66.38 & 45.67 & 52.73 & 56.31 & 55.47 & 34.36 & 63.90 & 64.84 & 65.30 & 51.65 & 60.20 & 39.17 \\
CFSMO & 57.41 & 39.82 & 50.30 & 54.57 & 39.99 & 29.55 & 62.38 & 66.57 & 63.29 & 52.73 & 50.55 & 31.69 \\
\bottomrule
\end{tabular}
}
\end{table*}

\begin{figure}[t]
\centering
\includegraphics[width=0.5\textwidth]{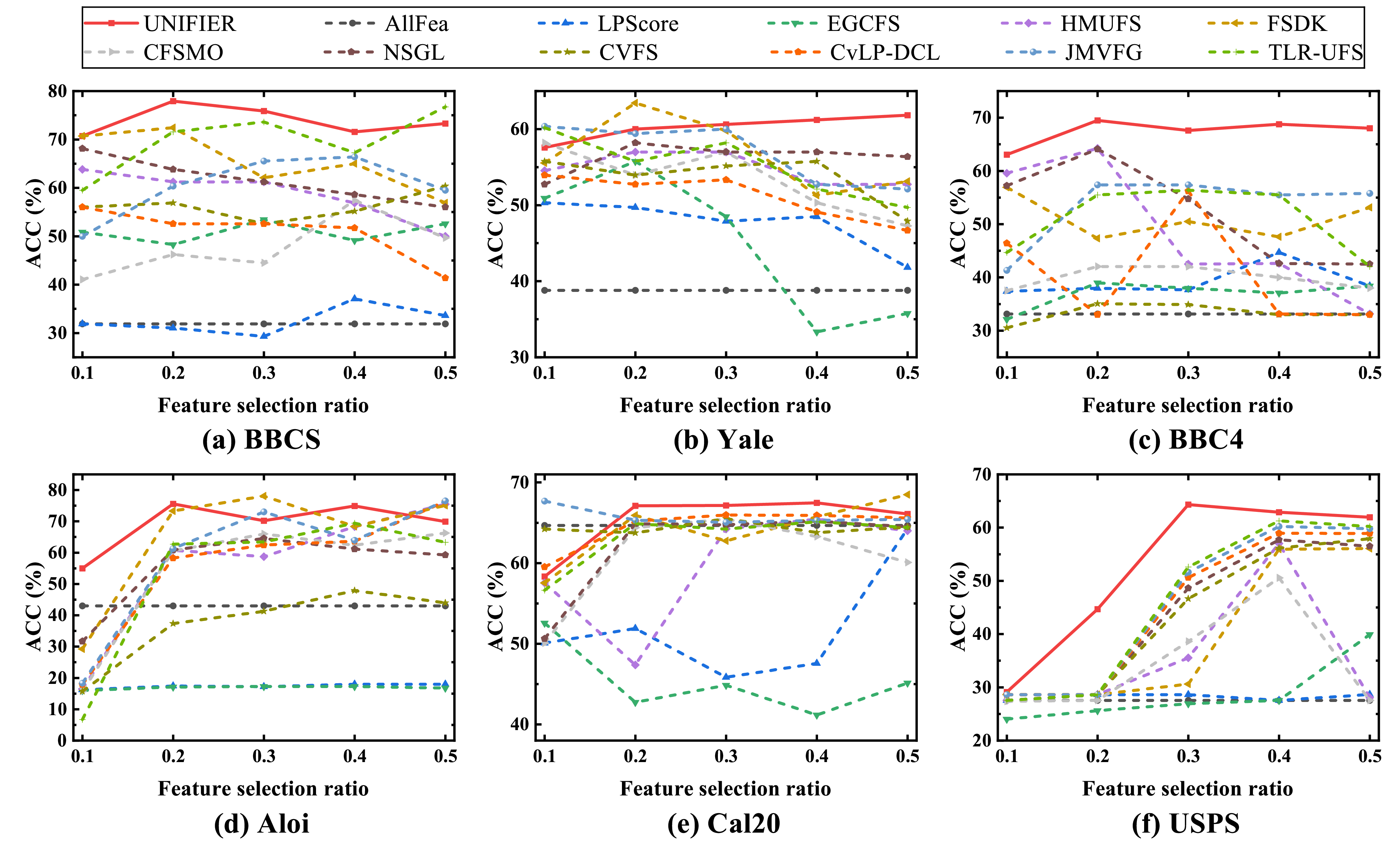}
\caption{ACC of different methods on six datasets under different feature selection ratios.}\label{ACC-missing0.2}
\end{figure}

\begin{figure}[t]
\centering
\includegraphics[width=0.5\textwidth]{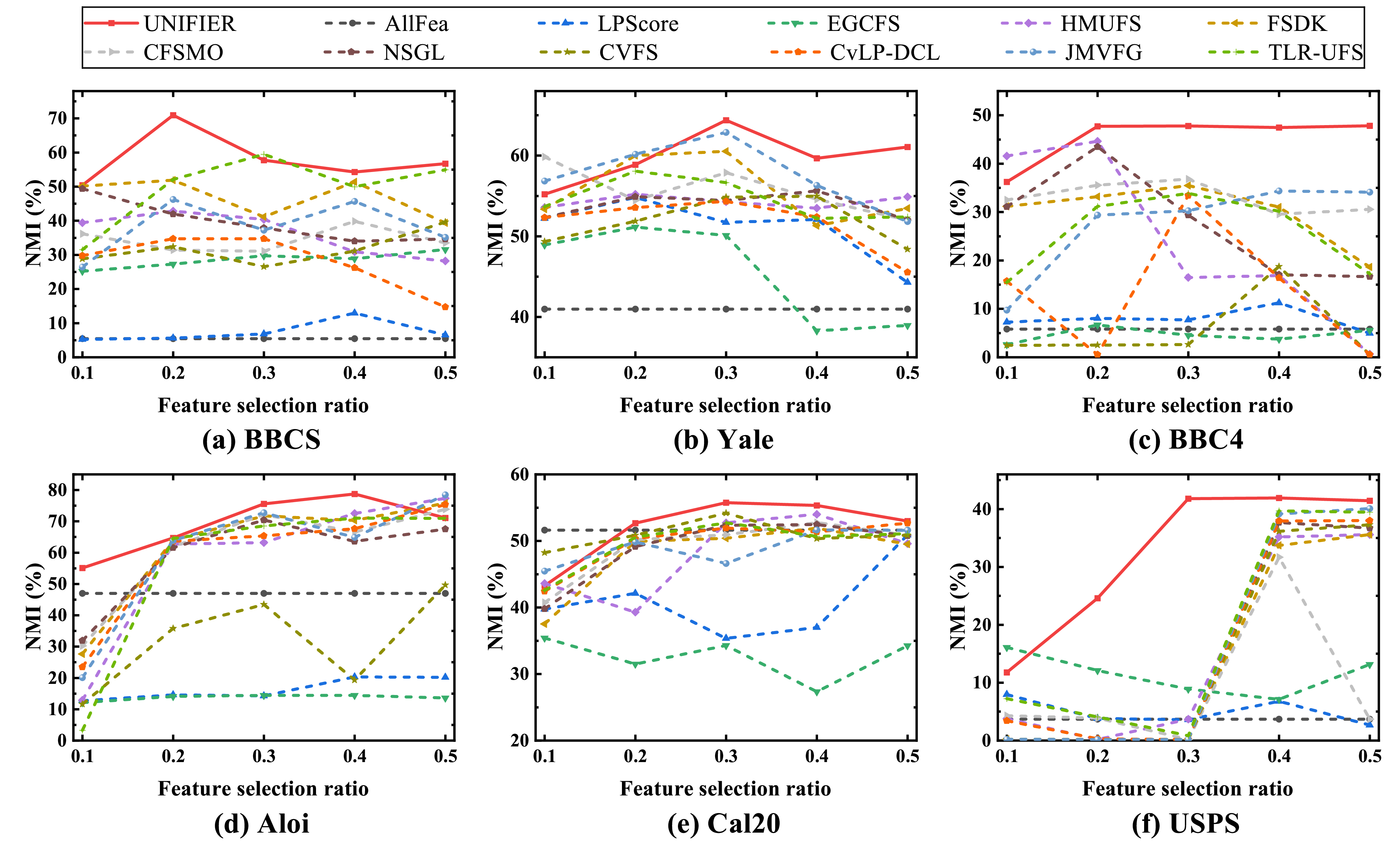}
\caption{NMI of different methods on six datasets under different feature selection ratios.}\label{NMI-missing0.2}
\end{figure}

\begin{figure}[t]
\centering
\includegraphics[width=0.5\textwidth]{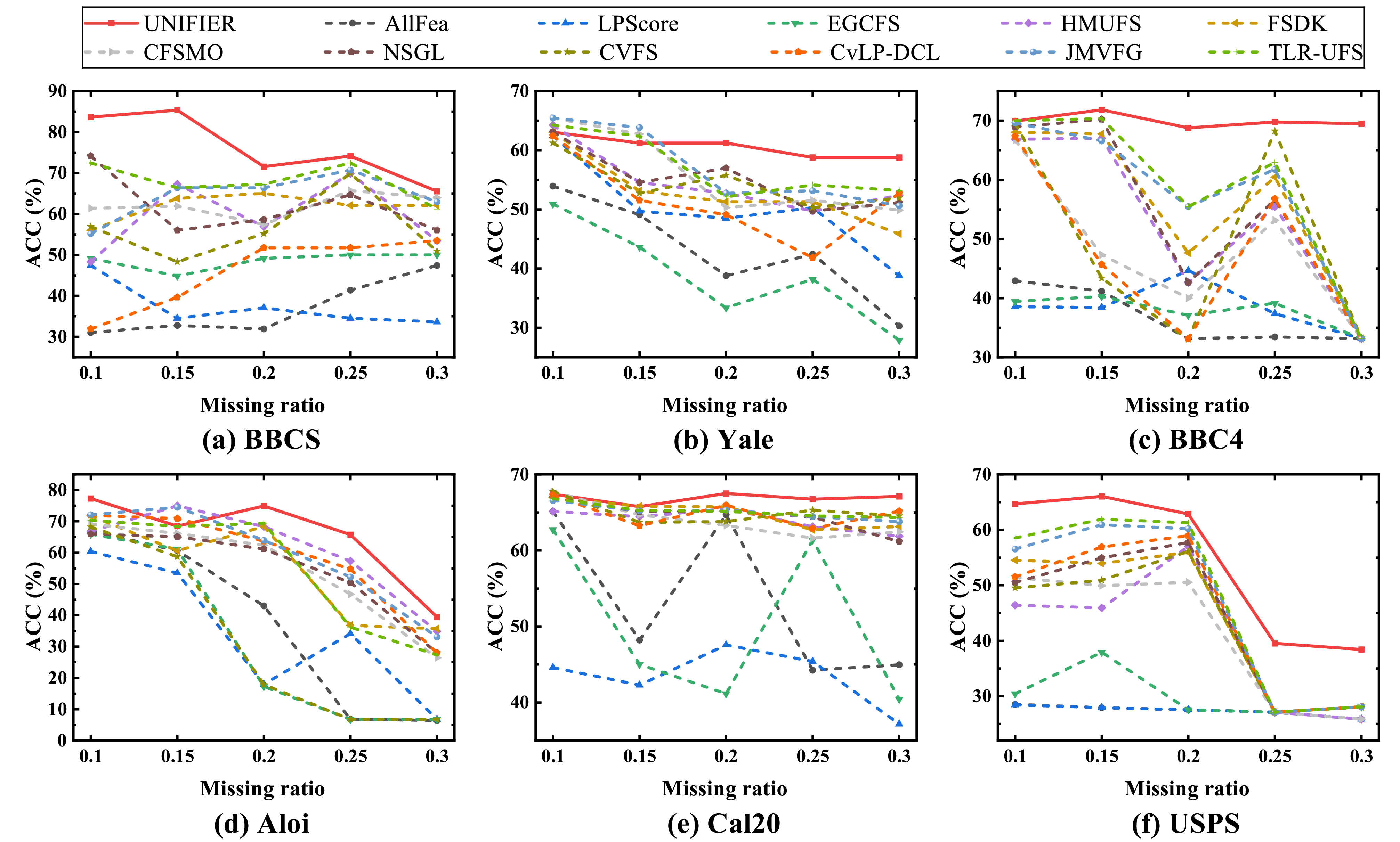}
\caption{ACC of different methods on six  datasets under different  missing rate.}\label{ACC-fs0.4}
\end{figure}

\begin{figure}[t]
\centering
\includegraphics[width=0.5\textwidth]{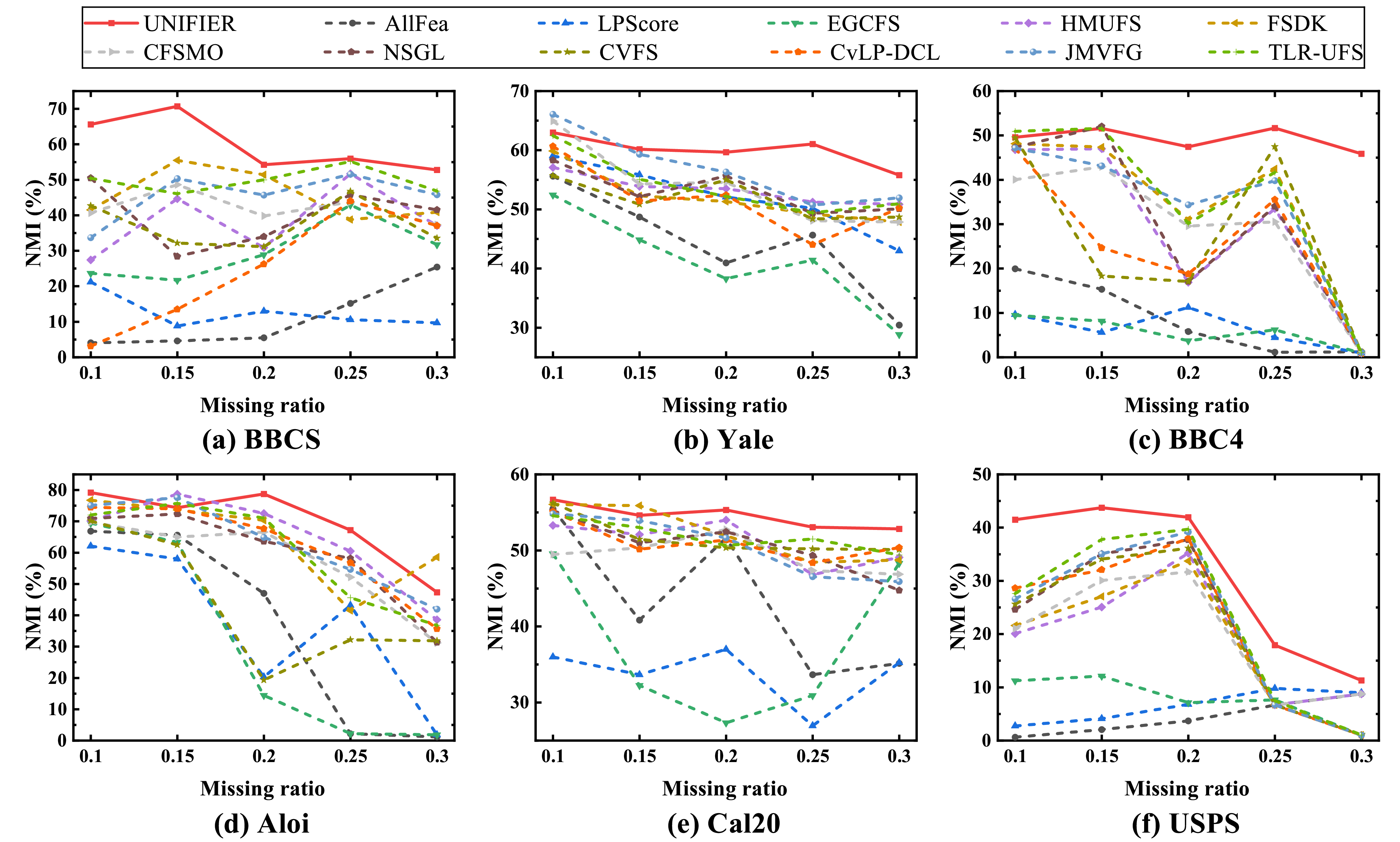}
\caption{NMI of different methods on six  datasets under different  missing rate.}\label{NMI-fs0.4}
\end{figure}

\textbf{Comparison methods.}
We compare UNIFIER with the following SOTA methods:

1) \textbf{AllFea} utilizes all the original features.

2) \textbf{LPscore}~\cite{LapScore} measures the importance of each feature by considering the locality-preserving power.

3) \textbf{EGCFS}~\cite{EGCFS} embeds between-class scatter matrix maximization into adaptive graph learning.

4) \textbf{NSGL}~\cite{NSGL} simultaneously learns the similarity graph and pseudo labels to select features.

5) \textbf{HMUFS}~\cite{HMUFS} utilizes an indicator matrix to exclude unobserved data from feature selection.

6) \textbf{FSDK}~\cite{FSDK} proposes a fast sparse discriminative K-means method for feature selection.

7) \textbf{CVFS}~\cite{CVFS} combines Hilbert-Schmidt Independence Criterion to perform cross-view feature selection.

8) \textbf{CvLP-DCL}~\cite{CvLP-DCL} learns cross-view similarity graphs for feature selection.

9) \textbf{TLR-MFS}~\cite{TLR-MFS} uses  tensor low-rank regularization to learn similarity graphs and select features.

10) \textbf{JMVFG}~\cite{JMVFG} integrates feature selection and multi-view graph learning into a unified framework.

11) \textbf{CFSMO}~\cite{CFSMO} introduces multi-order similarity learning for the selection of relevant features.

Since most comparison methods cannot be directly utilized with incomplete data, we fill the missing data with the mean values of features before employing these methods. To ensure a fair comparison, we tune the parameters of all methods using a grid search strategy and report the best performance. The parameters  $\alpha^{(v)}$ and $\lambda$ of our method are set within the range of $\{10^{-3}, 10^{-2}, 10^{-1}, 1, 10, 10^2, 10^3\}$. To simplify, the weight parameters for each view are set to be the same and the scale parameter $\gamma_{v}$ of Geman-McClure function is set to $1$. As determining the optimal number of selected features is still a challenging problem, we vary the proportion of selected features from $\{0.1, 0.2, 0.3, 0.4, 0.5\}$. We adopt a commonly used approach to assess MUFS~\cite{CvLP-DCL,R.Zhang2019}, employing clustering performance to evaluate the quality of selected features. In this paper, we run the graph-based multi-view clustering algorithm (GMC)~\cite{GMC} 30 times on the selected features and  use two widely recognized metrics, namely clustering accuracy (ACC) and normalized mutual information (NMI), to measure performance. As GMC is not sensitive to initialization, we only present the average results and omit the standard deviation.

\subsection{Performance Comparison}
Table 2 summarizes the performance of UNIFIER and other compared methods on six benchmark datasets, where we achieve the best ACC and NMI in all cases. As can be seen, on  two text datasets BBCS and BBC4, UNIFIER outperforms other competitors with average improvements of 21.91\% and 24.89\% in terms of ACC and NMI, respectively. For two object image datasets Aloi and Cal20, UNIFIER achieves average improvements of 14.08\% and 16.64\% in ACC and NMI, respectively. And in terms of ACC and NMI, the average improvements of UNIFIER on the handwritten image dataset USPS are 13.78\% and 13.85\%, respectively. Additionally, for the face image dataset Yale, UNIFIER continues to outperform other compared methods, respectively showing average improvements of 11.97\% and 8.55\% in ACC and NMI. 

Furthermore, to comprehensively validate the effectiveness of UNIFIER, we also present the results of all methods across various feature selection ratios and missing ratios. Figs. \ref{ACC-missing0.2} and \ref{NMI-missing0.2} respectively show the ACC and NMI values of comparative methods with different ratios of selected features, while the missing ratio is fixed at 0.2. It can be observed that UNIFIER yields the best performance in most cases compared with other methods when feature selection ratio varies from 0.1 to 0.5. Besides, Figs. \ref{ACC-fs0.4} and \ref{NMI-fs0.4} show the results of ACC and NMI with varying missing ratios while feature selection ratio is fixed at 0.4, respectively. As observed, UNIFIER consistently outperforms other comparison methods in most situations. The superior performance of the proposed method is attributed to the integration of feature selection, missing view imputation, and exploration of local structures within both sample and feature spaces in a unified learning framework, where these three components mutually enhance each other. Additionally, the dynamic sample quality assessment module in UNIFIER helps mitigate the impact of unreliable restored data and outliers.

\subsection{Parameters Sensitivity and Convergence}
\begin{figure}[t]
\centering
\includegraphics[width=0.48\textwidth]{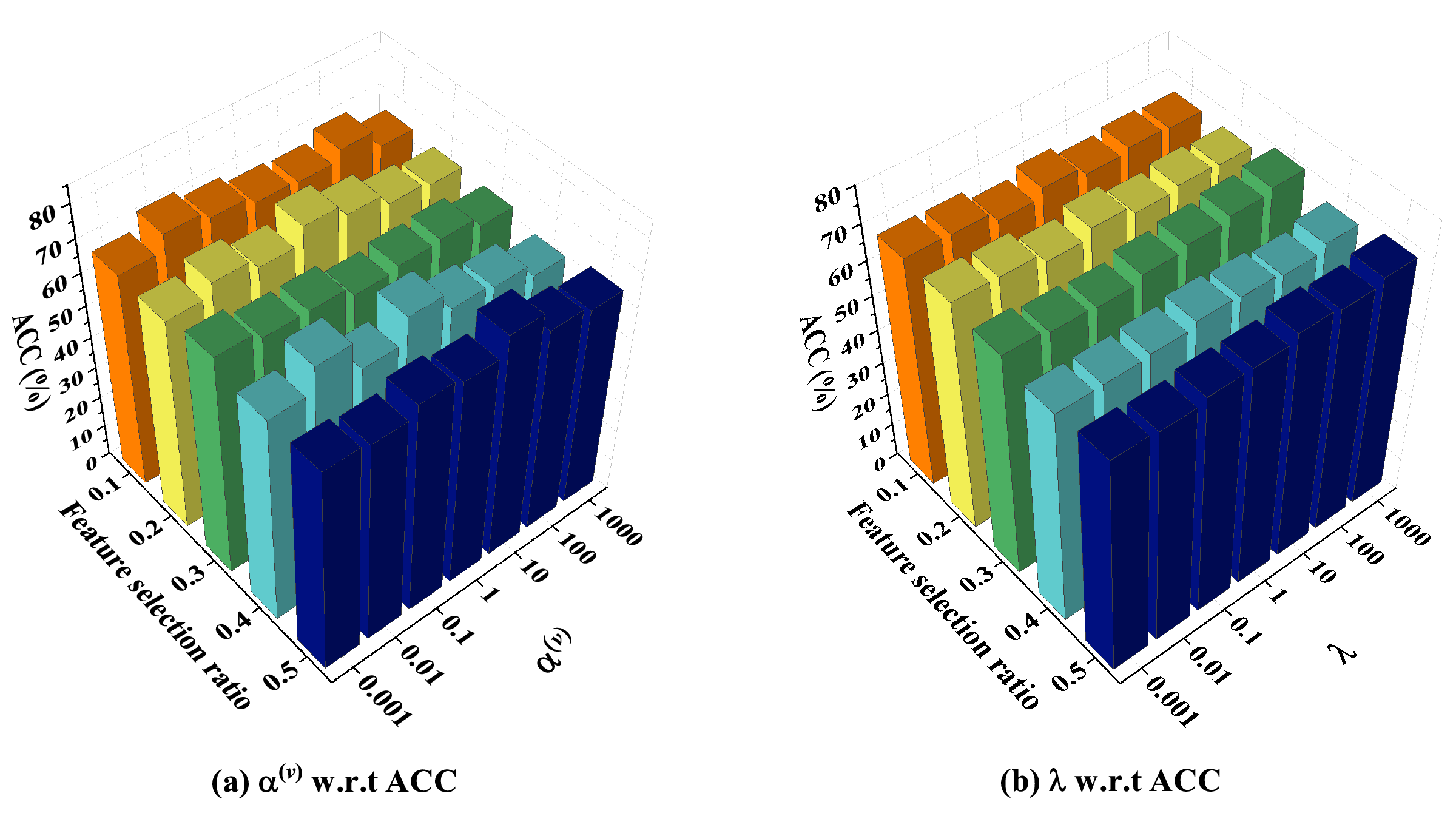}
\caption{ACC of UNIFIER with varying parameters $\alpha^{(v)}$, $\lambda$ and feature selection ratios on BBCS dataset.}\label{sensitivity}
\end{figure}

The objective function in Eq.(\ref{3.3.3}) involves three parameters: $\alpha^{(v)}$, $\lambda$, and $\xi_{v}$. The regularization parameter $\xi_{v}$ is automatically determined during the process of solving $\mathbf{S}^{(v)}$ in Eq.(\ref{4.14}). Hence, we only investigate the performance variation of our method at different values of $\alpha^{(v)}$ and $\lambda$. Fig.~\ref{sensitivity} shows ACC results on BBCS dataset, with one of $\alpha^{(v)}$ or $\lambda$ fixed, while the other varies across the range $\{10^{-3}, 10^{-2}, 10^{-1}, 1, 10^{1}, 10^{2}, 10^{3}\}$. We can observe that ACC shows a slight fluctuation w.r.t. $\alpha^{(v)}$ and remains relatively stable w.r.t. $\lambda$. This indicates that the parameter ranges of the proposed method can be readily tuned to attain satisfactory performance.

\begin{figure}[t]
	\centering
	\includegraphics[width=0.48\textwidth]{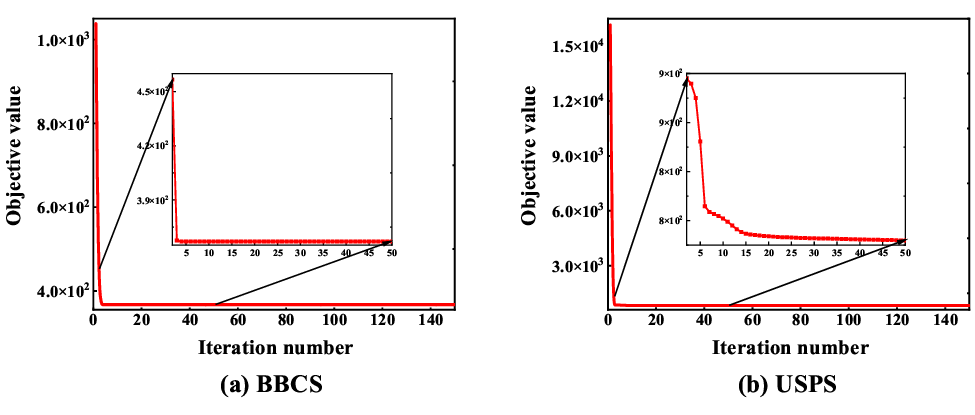}
	\caption{Object values of UNIFIER with different numbers of iterations on BBCS and USPS datasets.}\label{convergence}
\end{figure}

Fig.~\ref{convergence} illustrates the variation in the objective values of Eq.(\ref{3.3.3}) across different iterations on the BBCS and USPS datasets. It is evident that the proposed optimization algorithm converges rapidly within 20 iterations.

\subsection{Ablation Study}
In this section, we conduct ablation experiments to show the significance of the component terms in the proposed UNIFIER. Two variants of UNIFIER are conducted as follows:

(1) UNIFIER-\uppercase\expandafter{\romannumeral1} (without the bi-level collaborative missing view completion module):
\begin{equation}
\begin{aligned}
&\min_{\varTheta} \!\!\sum_{v=1}^{V} \!\!\bigg[ \alpha^{(v)} \!\sum_{i=1}^{n} \!\left(\!e_{i}^{(v)}\| \mathbf{x}_{i \cdot}^{(v)} \!-\! \mathbf{x}_{i \cdot}^{(v)}\mathbf{W}^{(v)} \|_{2}^{2} \!+\! \gamma(\sqrt{\mathbf{e}^{(v)}}\!-\!1)^{2}\right)\\
& + \frac{1}{2}\sum_{i,j=1}^n \| \mathbf{x}_{i \cdot}^{(v)} \mathbf{W}^{(v)} -\mathbf{x}_{j \cdot}^{(v)}\mathbf{W}^{(v)} \|_{2}^{2}s_{ij}^{(v)} + \xi_{v}\|\mathbf{S}^{(v)}\|_{F}^{2}\bigg]\\
\end{aligned}
\end{equation}
In UNIFIER-\uppercase\expandafter{\romannumeral1}, the missing samples in $\mathbf{X}^{(v)}$ are first filled with the means of the features, followed by the same feature selection process as in UNIFIER.

(2) UNIFIER-\uppercase\expandafter{\romannumeral2} (without the dynamic sample quality assessment module):
\begin{equation}
\begin{aligned}
&\min_{\varTheta} \sum_{v=1}^{V} \bigg[ \alpha^{(v)} \sum_{i=1}^{n} \| \widetilde{\mathbf{x}}_{i \cdot}^{(v)} - \widetilde{\mathbf{x}}_{i \cdot}^{(v)}\mathbf{W}^{(v)} \|_{2}^{2}+\lambda\|\mathbf{W}^{(v)}\|_{2,1}\\
& + \frac{1}{2}\sum_{i,j=1}^n \| \widetilde{\mathbf{x}}_{i \cdot}^{(v)} \mathbf{W}^{(v)} -\widetilde{\mathbf{x}}_{j \cdot}^{(v)}\mathbf{W}^{(v)} \|_{2}^{2}s_{ij}^{(v)} + \xi_{v}\|\mathbf{S}^{(v)}\|_{F}^{2}\bigg]\\
\end{aligned}
\end{equation}

Table~\ref{Ablation Study}  shows the ablation experiment results on six datasets. We can observe a significant decrease in the performance of UNIFIER-\uppercase\expandafter{\romannumeral1} compared to UNIFIER in terms of ACC and NMI. This underscores the benefit of integrating feature selection with missing view completion to improve  performance. Furthermore, UNIFIER outperforms UNIFIER-\uppercase\expandafter{\romannumeral2}, highlighting the effectiveness of the dynamic sample quality assessment module in mitigating the influence of unreliable restored data and outliers.

\section{Conclusions}\label{sec:Conclusions}
In this paper, we have proposed a novel MUFS method for incomplete multi-view data, addressing the limitation of current methods that treat missing data imputation and feature selection as two separate processes. The proposed method UNIFIER integrates feature selection, missing view completion, and the exploration of both local sample and feature structures into a unified learning framework. Furthermore, UNIFIER can automatically assign lower weights to low-quality samples, which is beneficial for alleviating the impact of unreliable restored data and outliers. Experiment results on benchmark datasets demonstrated the superior performance of UNIFIER compared with SOTA methods.

\begin{table}[!htbp]
\centering
\caption{Means (\%) of ACC and NMI for different variants of UNIFIER on six multi-view datasets.}\label{Ablation Study}
\vspace{-0.2cm}
\footnotesize
\renewcommand\tabcolsep{6.5pt} 
\begin{tabular}{lllllll}
\toprule[1pt]
\multirow{2}*{Datasets} & \multicolumn{2}{c}{UNIFIER}&\multicolumn{2}{c}{UNIFIER-\uppercase\expandafter{\romannumeral1}} &\multicolumn{2}{c}{UNIFIER-\uppercase\expandafter{\romannumeral2}} \\
\cmidrule(r){2-3} \cmidrule(r){4-5} \cmidrule(r){6-7}
~&ACC&NMI&ACC&NMI&ACC&NMI\\
\hline
\addlinespace[2pt]
BBCS  &\textbf{71.55}  & \textbf{54.26} &63.79  &42.90   & 58.62 &42.17 \\
Yale   &\textbf{61.21}  &\textbf{59.66}  &53.94  &55.37   &59.01  &57.53 \\
BBC4   &\textbf{68.76}  &\textbf{47.45}  &42.77  &17.48   &65.22  &44.03 \\
Aloi   &\textbf{74.91}  &\textbf{78.74}  &42.37  &47.53   &72.98  &73.51 \\
Cal20   &\textbf{67.48}  &\textbf{55.32}  &63.29  &48.67   &65.56  &53.21 \\
USPS   &\textbf{62.87}  &\textbf{41.93}  &58.98  &38.52   &58.24  &36.28 \\
\bottomrule
\end{tabular}
\end{table}

\end{document}